\documentclass{Interspeech}

\interspeechcameraready

\title{Suicide Risk Assessment Using Multimodal Speech Features: A Study on the SW1 Challenge Dataset}

\author[affiliation={1,2}]{Ambre}{MARIE}
\author[affiliation={1,2}]{Ilias}{MAOUDJ}
\author[affiliation={1,3}]{Guillaume}{DARDENNE}
\author[affiliation={1,3}]{Gwenolé}{QUELLEC}

\affiliation{}{LaTIM}{UMR INSERM 1101, Brest, France}
\affiliation{}{Université de Bretagne Occidentale}{Brest, France}
\affiliation{}{INSERM}{France}

\email{ambre.marie@univ-brest.fr, ilias.maoudj@univ-brest.fr, guillaume.dardenne@inserm.fr, gwenole.quellec@inserm.fr}
\keywords{suicide risk assessment, multimodal learning, speech embeddings, linguistic analysis}

\usepackage{comment}

\begin{document}

\maketitle{}

\begin{abstract}
The 1st SpeechWellness Challenge conveys the need for speech-based suicide risk assessment in adolescents. This study investigates a multimodal approach for this challenge, integrating automatic transcription with WhisperX, linguistic embeddings from Chinese RoBERTa, and audio embeddings from WavLM. Additionally, handcrafted acoustic features -including MFCCs, spectral contrast, and pitch-related statistics- were incorporated. We explored three fusion strategies: early concatenation, modality-specific processing, and weighted attention with mixup regularization. Results show that weighted attention provided the best generalization, achieving 69\% accuracy on the development set, though a performance gap between development and test sets highlights generalization challenges. Our findings, strictly tied to the MINI-KID framework, emphasize the importance of refining embedding representations and fusion mechanisms to enhance classification reliability.
\end{abstract}

\section{Introduction}

Suicide is recognized as a significant global health issue, ranking among the leading causes of death in adolescents \cite{wasserman_suicide_2021, Hua2023Suicide}. Early identification of individuals at risk is critical for timely intervention, yet reliable and accessible risk assessment remains a persistent challenge. Traditional suicide risk (SR) assessments are based primarily on clinical interviews and self-report questionnaires, both of which have significant limitations. Self-report assessments often lead to misclassification of suicidal behaviors, requiring follow-up questions to improve precision \cite{Hom2016Limitations}. Clinical interviews, while more exhaustive, require substantial human expertise and resources, making them difficult to scale \cite{Schechter2021The}. Additionally, the phrasing of interview questions can introduce bias, as negatively framed questions may lead patients to under-report suicidal ideation \cite{McCabe2017How}.

Speech analysis has been studied for its potential in classifying and predicting depression and suicidality, offering objective markers supporting clinical assessments \cite{Cummins2015A}. On the linguistic level, suicidal individuals tend to use more absolutist words and exhibit dichotomous thinking, characterized by polarized and negative linguistic structures \cite{Fekete2024Content,Fekete2018Linguistic}. On the paralinguistic level, suicidal speech often demonstrates altered prosodic characteristics, including changes in pitch, rhythm, and speech fluency \cite{Scherer2013Investigating}. By analyzing linguistic and paralinguistic markers, speech-based systems can provide objective tools for SR evaluation, helping clinicians identify individuals at risk \cite{Yünden2024Examination,Belouali2020Acoustic}. 

In response to this need, the 1st SpeechWellness Challenge was launched as part of Interspeech 2025. This challenge focuses on speech-based SR classification, providing a dataset of speech recordings from anonymized Chinese adolescents, with 50\% identified as having SR based on validated psychological scales.

In this work, we present a multimodal approach for assessing SR, integrating automatic transcription, text embeddings, audio embeddings, and acoustic feature extraction. We explore different fusion strategies for combining these modalities and evaluate their effectiveness. Our results provide insights into the strengths and limitations of multimodal learning for speech-based SR classification, and highlight areas for future research and improvement.

\section{Materials and methods}

Each submission differs in its choice of embeddings and fusion strategies: \textbf{(1)} Uses only audio (WavLM) and text (RoBERTa) embeddings, with an early fusion approach; \textbf{(2)} Introduces acoustic features (MFCCs) alongside audio and text, with a modality-specific fusion strategy; \textbf{(3)} Expands acoustic embeddings by incorporating spectral and pitch features, while using a weighted attention fusion mechanism.

\begin{figure}[t]
    \centering
    \includegraphics[width=\columnwidth]{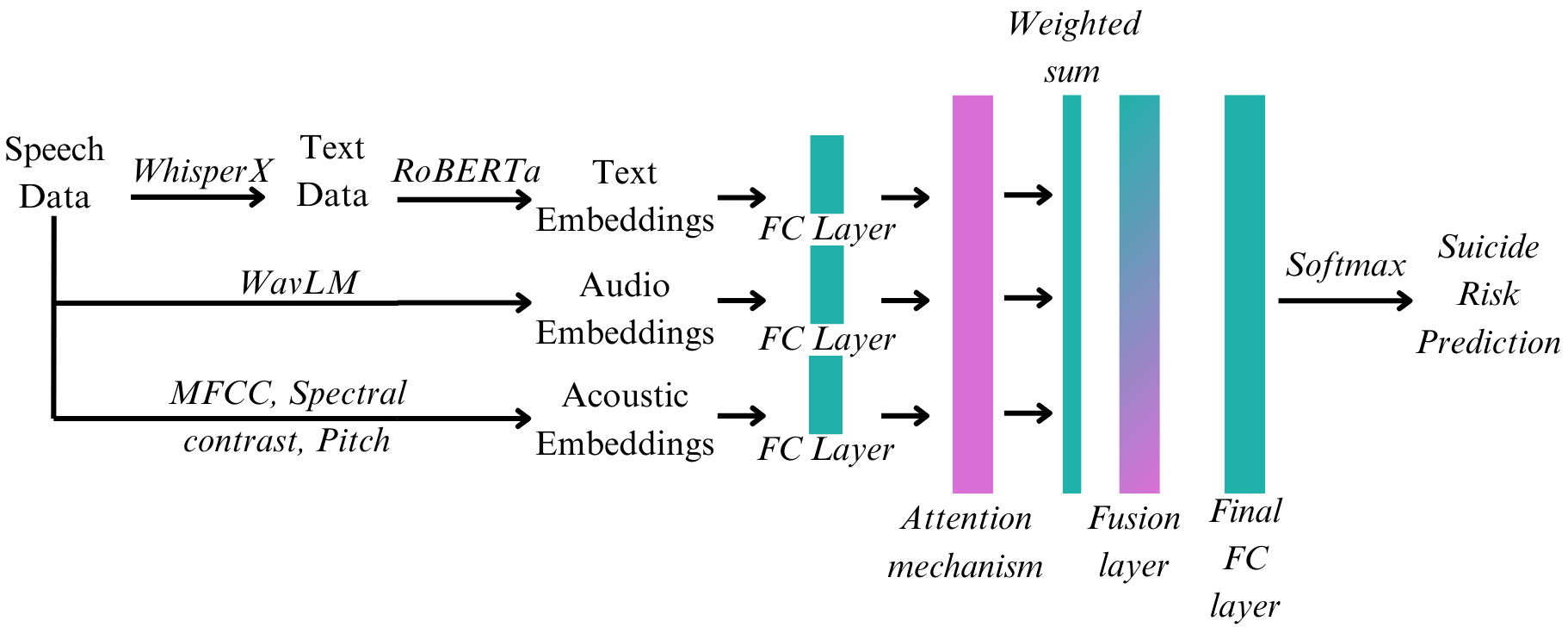}
    \caption{Architecture of the Submission 3 classification pipeline.}
    \label{fig:pipeline}
\end{figure}

Figure~\ref{fig:pipeline} provides an overview of the final processing pipeline, including transcription, embedding extraction, and fusion for classification.

\subsection{Dataset}

The dataset used in this study is the SW1 Challenge Dataset, provided as part of the 1st SpeechWellness Challenge at Interspeech 2025 \cite{wu20251stspeechwellnesschallengedetecting}. It consists of speech recordings from 600 anonymized Chinese adolescents aged 10 to 18 years. Participants were categorized into two groups: 300 subjects diagnosed with SR and 300 non-risk subjects, based on clinical evaluations using the Mini International Neuropsychiatric Interview for Children and Adolescents (MINI-KID).

Speech data was collected through interview tasks designed to capture linguistic and paralinguistic aspects of speech. Participants completed three tasks: an emotional regulation discussion, a standardized passage reading, and an image-based expression description.

The dataset included 400/100/100 samples in the training (train), development (dev) and test sets respectively. The sets were balanced and the labels in the test set remain undisclosed, ensuring an unbiased final evaluation. In this study, we adhered to the original split of the dataset, using the train and dev sets for model development and relying on the test set for the final submission.
%For model training and evaluation, the dataset was divided into three subsets:
%\begin{itemize}
%\item Training set: 400 subjects (200 at-risk, 200 non-risk)
%\item Development set: 100 subjects (50 at-risk, 50 non-risk)
%\item Test set: 100 subjects (labels not provided)
%\end{itemize}

\subsection{Transcription}

To incorporate both linguistic and acoustic information into the classification process, the audio recordings were transcribed into Mandarin text using the WhisperX \cite{bain2022whisperx} large-v2 model. By transcribing speech into structured text, we aimed to extract linguistic patterns that complement acoustic features. Previous studies have shown that multimodal models, which combine text and audio modalities, can outperform single-modal approaches by capturing a broader range of behavioral and emotional cues \cite{Ramírez-Cifuentes2020Detection}. 

\subsection{Embeddings}

To build a thorough representation of speech data, embeddings were extracted from audio, text, and acoustic features. Each type captures distinct but complementary aspects of the recordings. The following sections describe the extraction process for each type of embedding.

\subsubsection{Audio Embeddings}

To extract audio-based representations of speech, audio embeddings (AuE) were generated using WavLM \cite{Chen_2022}, a self-supervised model capturing high-level speech features relevant to emotion and classification.
Three AuE strategies were tested for SR detection.
In the first approach, WavLM Base+ was used to extract embeddings from non-overlapping 10-second chunks of speech. The embeddings were extracted from the last hidden state of WavLM, and the mean of all chunks was used as the final representation for each subject. 
The second approach employed WavLM Large, which has a greater number of parameters and a more expressive feature space. Windows with a 50\% overlap were also introduced to maintain continuity in speech and prevent information loss at chunk boundaries. Overlapping embeddings were aggregated via mean pooling.
The third approach allowed the model to capture short-term variations by reducing the chunk size to one second. While resulting in a higher dimensional representation, it provided a more detailed temporal view of speech features.
%\begin{enumerate}
%    \item \textbf{WavLM Base+}: Extracted embeddings from non-overlapping 10-second speech chunks. The final representation was obtained by averaging embeddings from the last hidden state of WavLM.
%    \item \textbf{WavLM Large}: Used a larger model with a more expressive feature space. Windows with 50\% overlap were introduced to maintain speech continuity and prevent information loss, with mean pooling applied to overlapping embeddings.
%    \item \textbf{Short-term variations}: Reduced chunk size to 1 second, increasing dimensionality but providing a more detailed temporal view of speech features.
%\end{enumerate}
Each of these embedding strategies offered different compromises between information granularity, computational efficiency, and context preservation. The objective was to  identify which approach provided the most discriminative features to improve model performance.

\subsubsection{Text Embeddings}

To encode the linguistic aspects of speech, text embeddings (TE) were extracted from the transcriptions using a transformer-based language model. These embeddings capture semantic and syntactic structures, allowing the model to differentiate between speech patterns associated with SR and those without. Given that the dataset consists of Mandarin transcriptions, pretrained Chinese RoBERTa \cite{liu2019robertarobustlyoptimizedbert} models were used to generate text representations.
For embedding extraction, two RoBERTa versions were tested: the base (efficient) and the large (richer representation) versions.
Each transcription was tokenized using the RoBERTa tokenizer, which splits the text into subword units. Then, the input was passed through the transformer model, generating contextualized word embeddings. Finally, the embeddings were aggregated into a single vector representation for each transcription using different pooling strategies. 
Three sentence embedding strategies were tested.
The first approach relied on extracting the [CLS] token embedding \cite{devlin2019bertpretrainingdeepbidirectional}, a widely used method for sentence-level representation in transformer-based models to provide a computationally efficient and compact representation. 
The second approach used mean pooling over all token embeddings. This ensured that the final representation incorporated information from the entire sentence rather than relying only on the [CLS] token. 
The last approach followed the same strategy, but used RoBERTa-large instead of the base model. Given the increased model size and depth, this method aimed to capture richer syntactic and semantic relationships in transcriptions.
These different embedding strategies influenced how linguistic information was captured and represented, balancing between compactness and expressiveness. Through evaluation, their classification performance was assessed, aiming to determine which approach best preserved the linguistic nuances relevant to SR detection.

\subsubsection{Acoustic Embeddings}

In the second and third versions of our work, acoustic embeddings (AcE) were introduced to complement AuE, incorporating detailed spectral and prosodic features. While AuE derived from WavLM contextual speech representations, AcE focused on handcrafted features, offering additional insights into voice articulation and pitch variations. 
In the second version, 40-dimensional MFCCs were computed, averaging each one over time to generate a fixed-length representation per utterance. This method captures articulation patterns, which may reflect changes in speech production associated with mental health conditions \cite{Suwannakhun2019Characterizing}.
For the third version, the acoustic representation was expended by including spectral contrast features and pitch estimates. Spectral contrast measures the energy difference between spectral peaks and valleys, providing information about phonation stability and articulation clarity. Additionally, fundamental frequency (f0)  and voiced probability metrics were extracted using librosa's pyin algorithm. These pitch-related features capture variations in intonation and prosody, which have been linked to suicidal speech characteristics \cite{Saavedra2020Association}.
In both versions, the extracted features were aggregated into a single vector representation per subject. The second version used only MFCC means, while the third version concatenated MFCCs, spectral contrast values, and pitch statistics.
By integrating AcE alongside WavLM AuE, the model combined self-supervised speech representations with manually engineered acoustic features. This approach aimed to enhance classification performance by leveraging both learned and interpretable speech descriptors, incorporating insights from both deep learning and traditional signal processing methods.

\subsection{Fusion Strategies}

To integrate the complementary information from text, audio, and AcE, different strategies were expolored. Each played a key role in determining how these diverse representations were combined and processed before being passed to the final classification model.

\subsubsection{Early Concatenation}

In the first version of our fusion pipeline, a simple early concatenation approach was adopted, where text, audio, and AcE were directly merged into a single feature vector before classification. This approach treated modalities equally without modeling interactions. A fully connected neural network was used to process the concatenated embeddings, mapping them to a two-class output representing SR presence or absence. The model was optimized using cross-entropy loss, and training was performed with batch gradient descent. This approach provided a straightforward baseline but lacked explicit attention to the varying contributions of each modality.

\subsubsection{Modality-Specific Processing with Attention}

The second approach introduced modality-specific processing before fusion by applying independent feedforward layers to each embedding type. Audio, text, and acoustic features were processed through separate fully connected layers with activation functions and normalization, as illustrated in Figure~\ref{fig:pipeline}. These transformed features were then concatenated and passed through an attention mechanism, which dynamically assigned weights to each modality. The final fused representation was passed to a classification layer. This approach aimed to improve classification performance by allowing the model to selectively focus on the most discriminative features across the modalities.

\subsubsection{Weighted Attention with Mixup Regularization}

The third version further refined the fusion process by incorporating a weighted attention mechanism alongside data augmentation using mixup regularization. The model learned modality importance weights via attention. These weights were applied to the transformed embeddings before fusion, allowing for a more flexible combination of information (Figure~\ref{fig:pipeline}). Additionally, mixup regularization was employed during training, where embeddings from different samples were interpolated to encourage generalization and prevent overfitting. This method used multiple training folds and ensemble learning strategies to improve robustness and stability in classification.

Through this progressive refinement of the fusion process, we sought to identify an approach that best used the complementary strengths of each modality, balancing interpretability, adaptability, and robustness in SR detection.

\section{Results}

%\begin{itemize}
%    \item \textbf{1:} Uses only audio (WavLM Base+) and text (RoBERTa-wwm-ext) embeddings, with an early fusion approach.
%    \item \textbf{2:} Introduces acoustic features (MFCCs) alongside audio and text, with a modality-specific fusion strategy.
%    \item \textbf{3:} Expands AcE by incorporating spectral and pitch features, while using a weighted attention fusion mechanism.
%\end{itemize}

\subsection{Overview of Performance Metrics}

Table~\ref{tab:results_overview} presents the main performance metrics for all three submissions. Each differs in its choice of embeddings, fusion strategies, and model variants. The table reports accuracy, F1-score, and AUROC on the dev set, along with accuracy on the held-out test set.

\begin{table}[th]
  \caption{Performance metrics on the dev and test sets. \\Sub.: Submission, Acc.: Accuracy.}
  \label{tab:results_overview}
  \centering
  \begin{tabular}{ccccc}
    \toprule
    \textbf{} & 
    \textbf{Dev Acc.} & \textbf{Dev F1} & \textbf{Dev AUROC} & \textbf{Test Acc.} \\
    \midrule
    Sub. 1
    & 0.70
    & 0.70
    & 0.70
    & 0.45 \\

    Sub. 2
    & 0.63
    & 0.63
    & 0.63
    & 0.53 \\

    Sub. 3
    & 0.69
    & 0.69
    & 0.69
    & 0.56 \\
    \bottomrule
  \end{tabular}
\end{table}

\subsection{Embedding Representation}

\begin{figure}[t]
    \centering
    \includegraphics[width=\columnwidth]{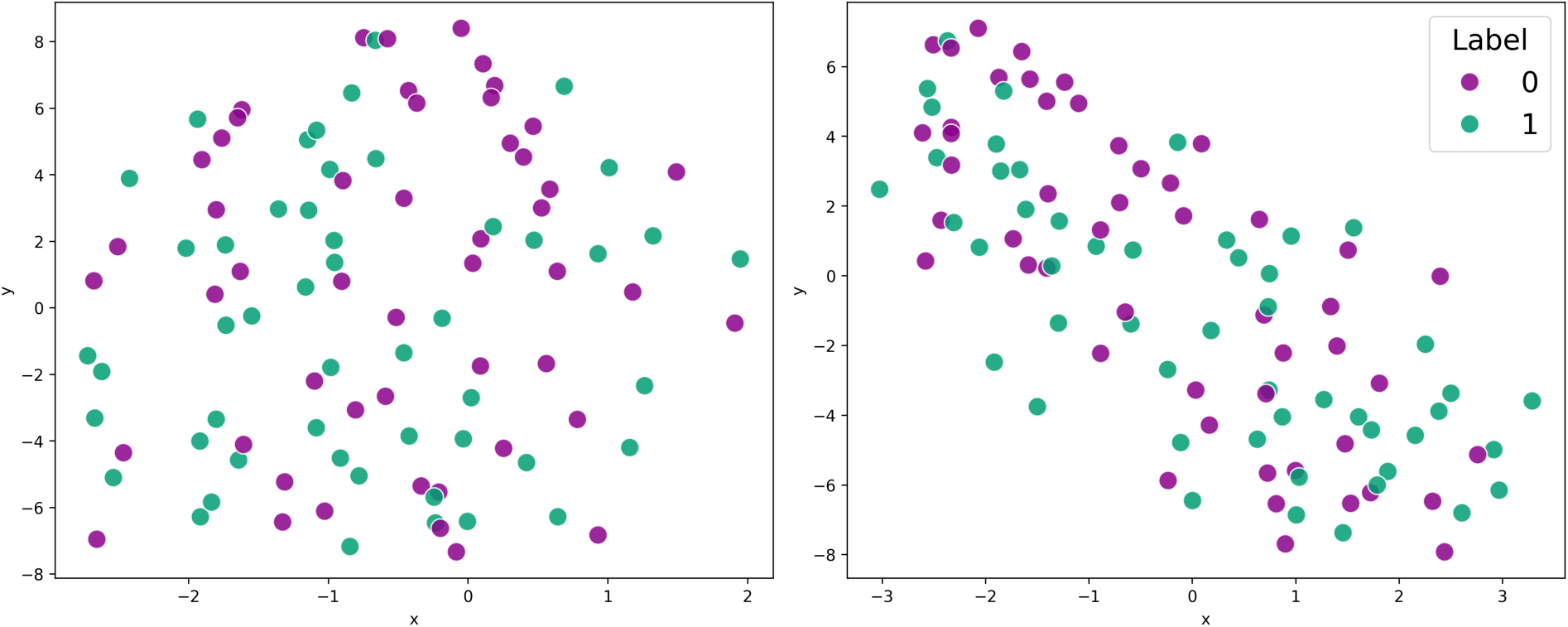}
    \caption{Visualization of AuE for Submission 2.
    Left: Raw AuE extracted from WavLM Large. 
    Right: Pre-logits AuE.\\ 
    Purple : label 0 (non-risk), Green : label 1 (at-risk).}
    \label{fig:audio_embeddings_example}
\end{figure}

To illustrate how the learned representations evolve from raw embeddings to the classifier's penultimate layer, we applied t-SNE to visualize both sets of embeddings. Figure~\ref{fig:audio_embeddings_example} depicts this transformation for Submission 2 applied to AuE. The generated figure presents a side-by-side comparison: the left panel shows raw embeddings extracted directly from pretrained models or handcrafted features, while the right panel shows the transformed embeddings after passing through the model's fusion and classification layers. 

\section{Discussion}

\subsection{Embeddings}

The choice of embeddings significantly influenced classification performance, with each modality contributing complementary information to the SR assessment task.

\subsubsection{Audio Embeddings}

AuE derived from WavLM provided a foundation for modeling speech characteristics. In the first version, WavLM Base+ was used with TE, which produced the highest dev accuracy of 70\%. However, this approach exhibited poor generalization, as indicated by a significant drop in test accuracy to 45\%. The model struggled to correctly classify test samples, particularly in distinguishing non-risk from at-risk cases.
The other methods employed WavLM Large, which provided a richer representation of speech features. Despite its increased capacity, the second version, which combined WavLM Large with additional AcE, achieved lower scores than the first one for the dev set, but improved test accuracy to 53\%. This result suggests that, while WavLM embeddings were effective in capturing speech patterns, they alone were insufficient for robust generalization. The t-SNE visualizations further support this observation, showing that raw AuE exhibited overlap between the two classes, while pre-logit embeddings demonstrated better separation after model training.

\subsubsection{Text Embeddings}

The use of TE from RoBERTa enabled the inclusion of linguistic markers potentially indicative of SR. In the first method, TE were paired with AuE in an early fusion model, producing the highest dev accuracy among all submissions. However, the test accuracy suffered considerably, suggesting that text-based features, while informative, may have led to overfitting when combined with audio in a simple concatenation framework.
Given the nature of the dataset, which includes both structured and unstructured speech tasks, linguistic markers alone may not have  been sufficiently discriminative, particularly in distinguishing between subtle variations in at-risk and non-risk speech patterns. In the first submission, a substantial misclassification rate was observed, particularly in false negatives, where at-risk individuals were often misclassified as non-risk. This suggests that TE, while capturing semantic information, require additional contextualization to be fully effective in this classification task.

\subsubsection{Acoustic Embeddings}

To enhance the model’s ability to capture paralinguistic features, the second and third methods introduced handcrafted AcE. Initially, 40-dimensional MFCCs were incorporated, which capture spectral envelope characteristics relevant to phonetic articulation and speech quality. While these features led to a drop in dev accuracy, they improved test accuracy compared to using no AcE. This suggests that while handcrafted features may not provide significant gains in training performance, they enhance generalization by capturing additional signal properties not explicitly modeled by deep learning representations.
The last version further extended acoustic features by incorporating spectral contrast and pitch-related embeddings, including fundamental frequency and voiced probability. This expansion increased test accuracy to 56\%, achieving the highest generalization performance among all versions. This more balanced classification indicated that the integration of spectral and prosodic characteristics helped mitigate biases present in previous approaches. This result aligns with prior research suggesting that prosodic cues, such as variations in pitch and articulation stability, can serve as reliable indicators of psychological distress \cite{Scherer2013Investigating2}.

\subsection{Fusion Strategies}

The fusion strategy was another key factor influencing model performance. Each method employed a different approach to integrating embeddings, ranging from early concatenation to more advanced attention-based mechanisms.

The early concatenation strategy was computationally simple and yielded the highest dev accuracy. However, the sharp decline in test accuracy highlights its major limitation: by treating all modalities equally, early concatenation failed to account for variations in feature importance across different samples. As a result, the model exhibited a strong bias, leading to an excessive number of misclassified cases, particularly in the test set. This suggests that a naive concatenation of embeddings, without explicit weighting or feature selection, may lead to overfitting, where the model relies too heavily on features that do not generalize well.

The Modality-Specific Processing with Attention approach addressed the limitations of early fusion by implementing separate processing layers for each modality before fusion. This aimed to enhance individual feature representations before combining them, allowing the model to extract more meaningful information from each embedding type. While this strategy led to a slight reduction in dev accuracy, the improvement in test accuracy suggests that it played a role in mitigating overfitting.
However, the results indicate that the model still struggled with classification, likely due to an imbalance in feature contributions. The lack of explicit weighting mechanisms may have resulted in suboptimal feature fusion, causing the model to rely too heavily on one modality over the others in certain cases. While modality-specific processing improved robustness, the results suggest that additional mechanisms were necessary to adjust the relative importance of each feature.

The weighted attention mechanism explicitly learned the contribution of each modality, allowing the model to emphasize the most informative embeddings while downweighting less relevant features. The addition of mixup regularization interpolated embeddings from different samples, thereby reducing the risk of overfitting to specific dataset biases.
This approach achieved the highest test accuracy of 56\%, demonstrating superior generalization than previous models. The t-SNE visualization of this method's embeddings supports this finding, as pre-logit embeddings exhibited a clearer distinction between at-risk and non-risk groups compared to raw embeddings. This suggests that learned fusion weights played an essential role in refining the final representation.

\subsection{Limitations and Future Perspectives}

The findings presented in this study are based on the scoring framework of the MINI-KID scale, which assesses current suicide risk as at risk or no risk. This classification reflects subjects’ immediate responses to the MINI-KID assessment and should not be interpreted as a prediction of future suicidal behaviour. Although the MINI-KID suicide module is widely recognised as a gold standard for assessing current suicide risk among adolescents, it has limitations. It relies heavily on self-reported data, which may lead to underreporting or misinterpretation of symptoms, and its fixed set of items may not fully capture the complex and dynamic nature of suicidal ideation and behaviour. Accordingly, the results reported herein are strictly confined to the context of this assessment.

Beyond the methodologies implemented in this study, we explored alternative approaches, such as fine-tuning a DeepSeek model with LoRA for few-shot learning on text transcriptions. While time constraints prevented full integration, this approach holds promise for capturing subtle linguistic markers in a more adaptable manner. Future work should focus on improving embedding representations to better distinguish between risk levels while developing fusion mechanisms that generalize well, balancing robustness with classification accuracy. Expanding the dataset to more diverse linguistic and cultural contexts would enhance model reliability. Furthermore, more reliable labeling, incorporating expert consensus or enhanced clinical metrics, would address the known limitations of self-reported data \cite{Verhoeven2017Agreement} and improve training stability. While challenges remain, this study contributes insights into multimodal speech analysis for suicide risk assessment and identifies key areas for improvement.

\section{Conclusion}

In this study, we explored a multimodal approach to suicide risk assessment using the SW1 Challenge dataset, integrating linguistic and paralinguistic features. Our findings suggest that combining self-supervised speech representations with handcrafted acoustic and linguistic features enhances classification performance. Among the fusion strategies tested, weighted attention demonstrated the highest generalization capability. However, while this approach showed promise, the performance gap between dev and test accuracy highlights the challenges of generalizing speech-based suicide risk assessment models beyond controlled datasets.  

\bibliographystyle{IEEEtran}
\bibliography{mybib}

\end{document}